\pdfoutput=1

\documentclass[11pt]{article}

\usepackage[final]{acl}

\usepackage{times}
\usepackage{latexsym}

\usepackage[T1]{fontenc}

\usepackage[utf8]{inputenc}

\usepackage{microtype}

\usepackage{inconsolata}

\usepackage{graphicx}

\usepackage{booktabs}
\usepackage{multirow}
\usepackage{makecell}
\usepackage{xspace}
\usepackage{amsmath}
\newcommand{\modelname}{KG$^2$RAG\xspace}
\usepackage{enumitem}

\title{Knowledge Graph-Guided Retrieval Augmented Generation}
    
\author{
Xiangrong Zhu$^\clubsuit$
\quad Yuexiang Xie$^\heartsuit$
\quad Yi Liu$^\clubsuit$
\quad Yaliang Li$^\heartsuit$
\quad Wei Hu$^\clubsuit$\\
$^\clubsuit$ State Key Laboratory for Novel Software Technology, Nanjing University, China \\
$^\heartsuit$ Alibaba Group \\
\texttt{\{xrzhu, yiliu07\}.nju@gmail.com, whu@nju.edu.cn} \\
\texttt{\{yuexiang.xyx, yaliang.li\}@alibaba-inc.com}
}

\begin{document}
\maketitle

\begin{abstract}
Retrieval-augmented generation (RAG) has emerged as a promising technology for addressing hallucination issues in the responses generated by large language models (LLMs). Existing studies on RAG primarily focus on applying semantic-based approaches to retrieve isolated relevant chunks, which ignore their intrinsic relationships.
In this paper, we propose a novel Knowledge Graph-Guided Retrieval Augmented Generation (\modelname) framework that utilizes knowledge graphs (KGs) to provide fact-level relationships between chunks, improving the diversity and coherence of the retrieved results. Specifically, after performing a semantic-based retrieval to provide seed chunks, \modelname employs a KG-guided chunk expansion process and a KG-based chunk organization process to deliver relevant and important knowledge in well-organized paragraphs.
Extensive experiments conducted on the HotpotQA dataset and its variants demonstrate the advantages of \modelname compared to existing RAG-based approaches, in terms of both response quality and retrieval quality. 
\end{abstract}

\section{Introduction}
\label{sect:intro}

Recently, large language models (LLMs)~\cite{li24llmsurvey,ren24llmsurvey,hugo23llama,tom20gpt} have achieved remarkable success across a broad range of real-world tasks, including question answering~\cite{sen23knowledge}, writing assistance~\cite{marco23cicero}, code generation~\cite{cheng24data}, and many others~\cite{jean23llmapplication,wu23autogen}. However, hallucinations~\cite{xu24hallucinationsurvey,liu24hallucinationsurvey} in the generated responses becomes a critical challenge, which often results from containing outdated information or lacking domain-specific knowledge.
Retrieval-augmented generation (RAG)~\cite{gao23ragsurvey,fan24ragsurvey} has emerged as a feasible solution to mitigate hallucinations by retrieving relevant knowledge from provided documents and incorporating it into the prompts of LLMs for response generation.

\begin{figure}[t]
\centering
\includegraphics[width=\linewidth]{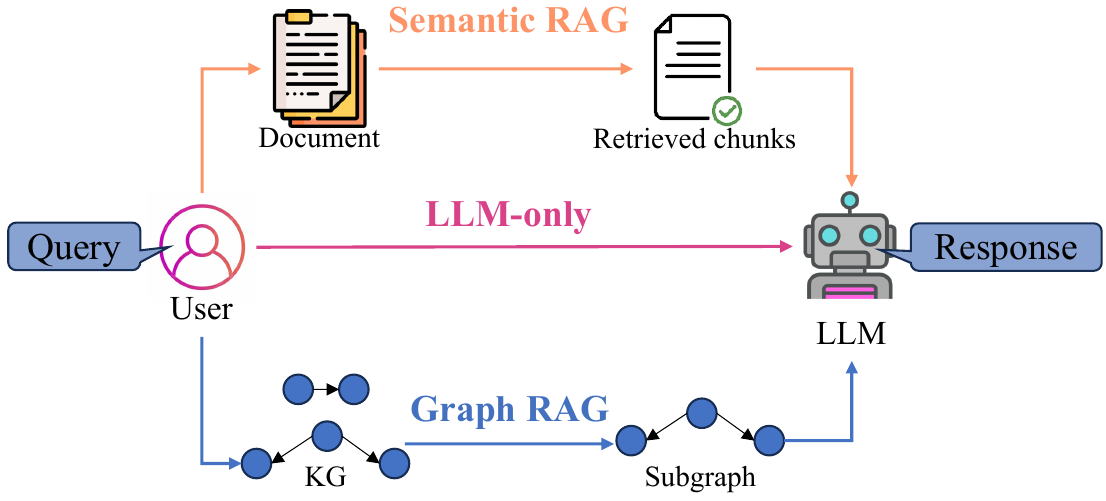}
\caption{A comparison among LLM-only, Semantic RAG, and Graph RAG paradigms.}
\label{fig:paradigms}
\end{figure}

Existing studies in RAG~\cite{patrick20rag,yu22rag,anupam23keyword,gao23ragsurvey,angelo24ratts}, as shown in Fig.~\ref{fig:paradigms}, employ keyword-based or semantic-based approaches to retrieve documents or chunks having the highest similarities to user queries. However, these retrieved chunks can be homogeneous and redundant, which fails to provide the intrinsic relationships among these chunks and cannot further activate the reasoning abilities of LLMs. Furthermore, the retrieved chunks are often directly concatenated in the order of their similarity scores and fed into LLMs as part of the prompts. Such a practice can lead to isolated pieces of information, limiting the utility of LLMs in generating comprehensive and reliable responses.

Knowledge graphs (KGs)~\cite{auer07dbpedia,ji22kgsurvey}, as structured abstractions of real-world entities and their relations, can be expected to effectively supplement existing semantic-based RAG approaches by integrating structured factual knowledge. 
Knowledge within a KG, represented in the form of triplets (\textit{head entity}, \textit{relation}, \textit{tail entity}), is naturally linked through overlapping entities.
A simplified workflow for utilizing KGs in RAG is shown in Fig.~\ref{fig:paradigms}, where relevant triplets are retrieved to augment the context for response generation in LLMs, providing fact-level relationships among chunks and highlighting important facts that may be missed by semantic-based approaches.

Shed light by such insights, in this paper, we propose a novel \textbf{K}nowledge \textbf{G}raph-\textbf{G}uided \textbf{R}etrieval \textbf{A}ugmented \textbf{G}eneration framework, called \modelname.
Specifically, we first perform chunking and KG-chunk association during the offline processing of the provided documents, establishing linkages between chunks and a specific KG to capture the fact-level relationships among these chunks. 
Based on the chunks and the KG, \modelname employs KG-enhanced chunk retrieval, which consists of a semantic-based retrieval and graph-guided expansion. The semantic-based retrieval prepares several seed chunks using embedding and ranking techniques~\cite{zach24nomic,li24mxbai}. These seed chunks are then used to extract a relevant subgraph from the association KG, onto which we can apply graph traversal algorithms to include the chunks containing overlapped or related entities and triplets. Such a design of graph-guided expansion provides a greater diversity of retrieved chunks and a comprehensive knowledge network.

After that, we incorporate a post-processing stage named KG-based context organization in \modelname. On one hand, the KG-based context organization serves as a filter to retain the most relevant information contained in the subgraph, thereby enhancing the informativeness of the retrieved chunks. On the other hand, it serves as an arranger to organize the chunks into internally coherent paragraphs with the knowledge graph as a skeleton. 
These semantically coherent and well-organized chunks are fed into the LLMs along with user queries for response generation.

We conduct a series of experiments on the widely-used HotpotQA~\cite{yang18hotpotqa} dataset and its newly constructed variants to mitigate the impacts of prior knowledge on LLMs. We adopt a distractor and a fullwiki setting, comparing \modelname with several RAG-based approaches. The experimental results demonstrate that \modelname consistently outperforms baselines in terms of both response quality and retrieval quality. Moreover, we conduct an ablation study to highlight the effectiveness of different modules in \modelname.
The constructed dataset and source code are released at \url{https://github.com/nju-websoft/KG2RAG} to further promote the development and application of KGs in RAG.

\begin{figure*}[t]
\centering
\includegraphics[width=0.98\linewidth]{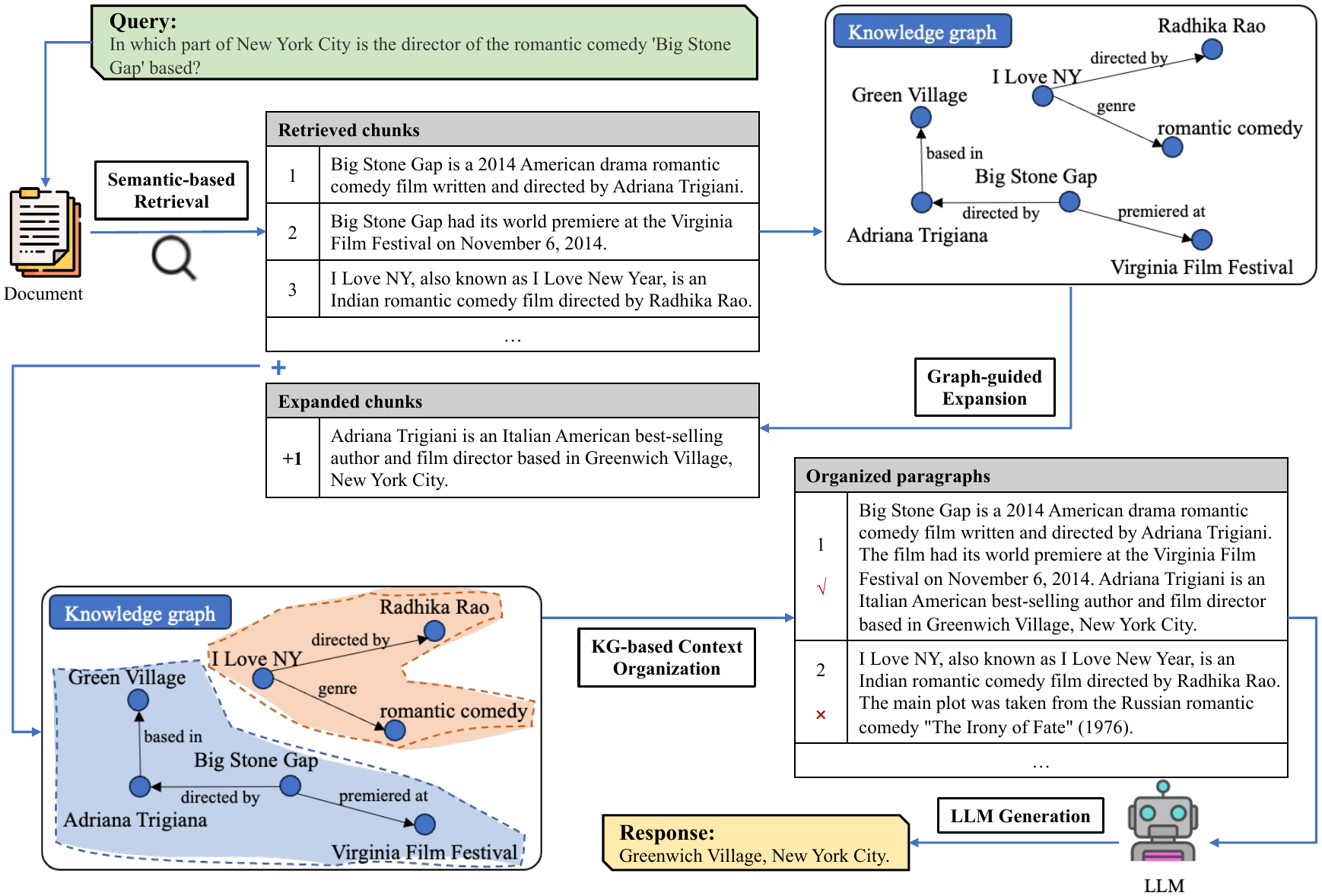}
\caption{Workflow of the proposed \modelname.}
\label{fig:pipeline}
\end{figure*}

\section{Methodology}
\label{sec:method}

An overview of the workflow of \modelname is illustrated in Fig.~\ref{fig:pipeline}. 
In the following subsections, we provide more details following the workflow of \modelname, including document offline processing (Sec.~\ref{subsec:offline}), KG-enhanced chunk retrieval (Sec.~\ref{subsec:retrieval}), and KG-based context organization (Sec.~\ref{subsec:context}).

\begin{figure}[t]
\centering
\includegraphics[width=\columnwidth]{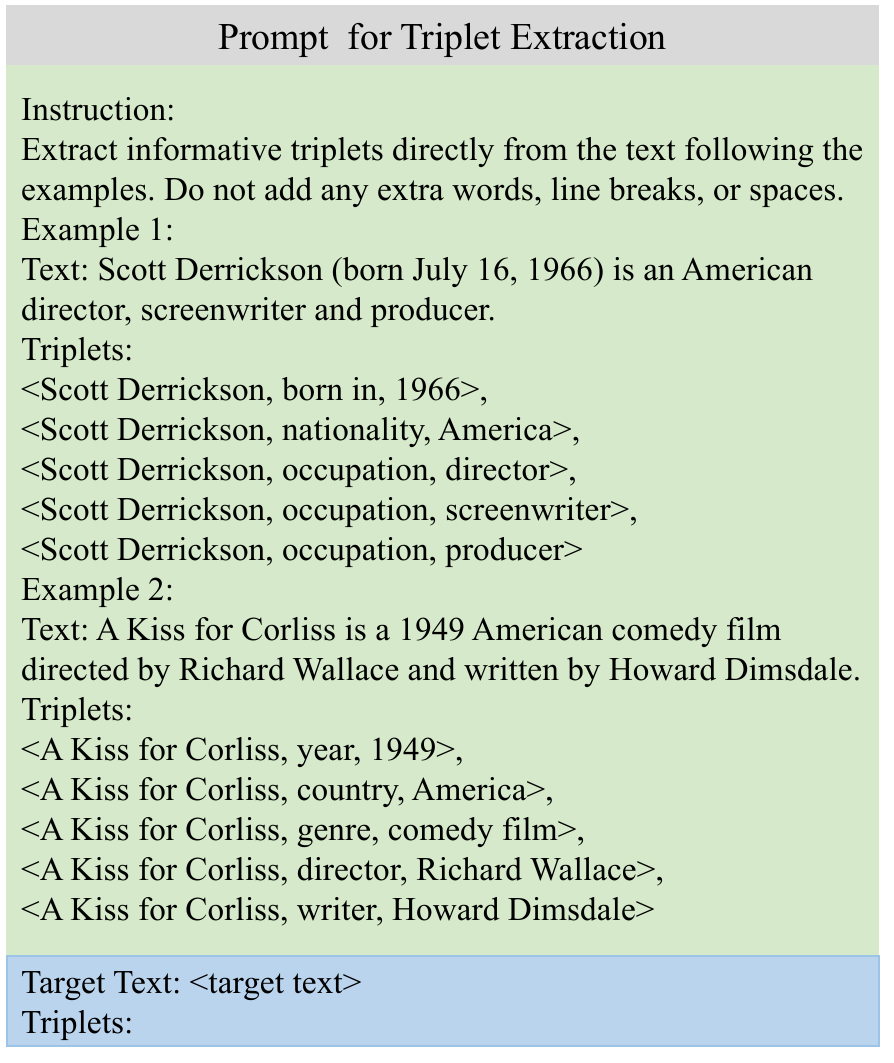}
\caption{The prompt for triplet extraction.}
\label{fig:extract_prompt}
\end{figure}

\subsection{Document Offline Processing}
\label{subsec:offline}
Following the existing studies in RAG~\cite{patrick20rag,gao23ragsurvey,fan24ragsurvey}, all documents are first split into $n$ chunks based on the structure of sentences and paragraphs given a predefined chunk size, which can be given as $\mathcal{D} = \{c_1,\ldots,c_n\}$.
These chunks can be further processed, for example, by adding relevant context~\cite{jiang23arag,matous24aragog}, extracting meta-information~\cite{laurent24meta} (e.g., title, abstract), and generating corresponding questions~\cite{ma23queryrewriting,wang24maferw}. 
Since these chunk-enhancing techniques are orthogonal to the proposed method in this paper, we recommend referring to the original paper for more details.
Hereafter, we continue to denote the processed chunks as $\mathcal{D} = \{c_1,\ldots,c_n\}$.

To capture the rich fact-level relationships among these chunks, we associate them with a KG,  which can be implemented via the following approaches. In cases where a KG is available, such as in WebQSP~\cite{tih16webqsp} and CWQ~\cite{alon18cwq}, the chunk-KG association can be performed through entity and relation recognition and linkage algorithms~\cite{zhao23improving,tian24generating}. 
Another approach involves directly extracting multiple entities and relations from the chunks to form subgraphs, which can be used to combine into a complete graph. 
In this paper, to avoid reliance on existing KGs, we adopt the latter approach, implementing it by providing appropriate prompts (refer to Fig.~\ref{fig:extract_prompt}) to LLMs.

After this process, we provide linkages between chunks and a specific KG, which can be given as
\begin{equation}
    \mathcal{G} = \{(h,r,t,c)\,|\,c\in \mathcal{D}\}, 
\end{equation}
where $h$, $r$, and $t$ denote the head entity, relation, and tail entity, respectively, and $c$ denotes the chunk that derives the triplets.
Note that the chunk-KG association process is query-independent, which implies that it can be performed offline, only needs to be constructed once for all documents, and supports incremental updates for new documents.
As the document offline processing aligns with what vanilla RAG does, \modelname naturally supports adding new documents to or removing documents from the existing knowledge base and KG efficiently.

\subsection{KG-enhanced Chunk Retrieval}
\label{subsec:retrieval}
Given the chunks $\mathcal{D}$ and the associated KG $\mathcal{G}$, the proposed \modelname suggests a two-stage retrieval process, including semantic-based retrieval and graph-guided expansion. 

\paragraph{Semantic-based Retrieval}
During the semantic-based retrieval process, the semantic similarities between a user query $q$ and all the chunks can be measured as
\begin{equation}
\mathcal{S} = \{s(q,c)\,|\,c\in\mathcal{D}\},
\label{eq:embedding_retrieval}
\end{equation}
where the similarity function $s(\cdot)$ employs an embedding model~\cite{zach24nomic,li24mxbai} to transfer the query and chunks into high-dimensional representations, followed by computing their cosine similarity.

The chunks with the top-$k$ highest similarities to the query are selected as the retrieved chunks, denoted by $\mathcal{D}_q$. These retrieved chunks can be integrated into the prompts as context and fed into LLMs for RAG. 
As discussed in Sec.~\ref{sect:intro}, relying solely on semantic-based retrieval may result in isolated chunks, missing crucial factual knowledge and the intrinsic connections among the chunks.
To tackle this, we regard the retrieved chunks $\mathcal{D}_q$ as seed chunks, and propose a graph-guided expansion process.

\paragraph{Graph-guided Expansion}
During communication and thinking processes, people often connect one event to others as these events involve the same entities, such as persons and places.
For example, \textit{Capitol Hill, Washington, D.C.} connects our impressions of \textit{Barack Obama}, \textit{Donald Trump}, and \textit{Joe Biden}, as they all delivered their presidential inaugural speeches there in 2013, 2017, and 2021, respectively.
Shed light by such insights, \modelname suggests linking one chunk to other chunks through the overlapping or connected entities that they contain for retrieved chunk expansion.

Specifically, given the retrieved chunks $\mathcal{D}_q\subseteq\mathcal{D}$ and the KG $\mathcal{G}=\{(h,r,t,c)\,|\,c\in\mathcal{D}\}$, we first get the relevant subgraph of $\mathcal{D}_q$ as follows:
\begin{equation}
\label{eq:subgraph}
\mathcal{G}_q^0 = \{(h,r,t,c)\,|\,c\in\mathcal{D}_q\}\subseteq\mathcal{G}.
\end{equation}

After that, we traverse the $m$-hop neighborhood of $\mathcal{G}_q$ to get the expanded subgraph $\mathcal{G}_q^m$, which can be given as
\begin{equation}
\label{eq:traverse}
\mathcal{G}_q^m = \text{traverse}(\mathcal{G},\mathcal{G}_q^0,m),
\end{equation}
where $\text{traverse}(\cdot)$ can be implemented with the breadth-first search (BFS) algorithm,
serving as a function that captures all entities in $\mathcal{G}_q^0$, corresponding $m$-hop neighboring entities, and all edges linking these entities to form an expanded subgraph.

Given the expanded subgraph $\mathcal{G}_q^m$, we can readout all the chunks associated with the graph (i.e., containing facts corresponding to the triplets in this graph) as follows:
\begin{equation}
\label{eq:expansion}
\mathcal{D}_q^m = \{c\,|\,(h,r,t,c)\in\mathcal{G}_m^q\}\subseteq\mathcal{D},
\end{equation}
where $\mathcal{D}_q^m$ is referred to as the expanded chunks.

\paragraph{Discussions}
Several semantic-based and context-based approaches can also achieve chunk expansion. For example, one can increase the value of $k$ in the aforementioned similarity-based retrieval process, or apply a context window expansion~\cite{jiang23arag} (i.e., when a chunk is retrieved, the chunks within the context window are also recalled together). Different from these approaches, the proposed graph-guided expansion gathers chunks that contain the same or related entities or triplets, without requiring these expanded chunks to have high semantic similarity to the query or to be located around the retrieved chunks. 
Such a design of graph-guided expansion helps prevent redundancy and excessive homogeneity among the retrieved and expanded chunks, leading to greater diversity and the development of a more comprehensive knowledge network.
We provide some empirical evidence to further confirm the effectiveness of the proposed graph-guided expansion in Sec.~\ref{sec:discussions}.

\subsection{KG-based Context Organization}
\label{subsec:context}
After the KG-enhanced chunk retrieval, \modelname incorporates a post-processing stage before response generation of LLMs, motivated by the following two considerations.

Firstly, the number of expanded chunks through the graph-guided expansion is tied to the triplets contained in the expanded subgraph, which can be too large, potentially exceeding the context length and introducing noise that may obscure helpful information.
Secondly, inspired by human reading habits and previous studies~\cite{li23promptsurvey,liu24lost}, providing semantically coherent and well-organized materials as context makes positive impacts on the understanding and generation performance of LLMs.
As a result, we propose a KG-based context organization module in \modelname, which serves as both a filter and an arranger to meet these requirements.

\paragraph{Serving as a Filter}
Specifically, we first calculate the semantic similarities between the expanded chunks with the user query, according to Eq.~\eqref{eq:embedding_retrieval}. Based on these similarities, the expanded subgraph $\mathcal{G}_q^m$ can be transformed into an undirected weighted graph as follows:
\begin{equation}
\label{eq:undirect_graph}
\begin{aligned}
\mathcal{U}_q^m=\{&(h\leftrightarrow t, \text{rel}:r, \text{src}:c, \text{weight}:s(q,c))\\
&\,|\,(h,r,t,c)\in\mathcal{G}_q^m\},
\end{aligned}
\end{equation}
where $h\leftrightarrow t$ represents an undirected edge, attached with the corresponding relation and the source chunk as meta information.
We reuse the semantic similarities calculated in Sec.~\ref{subsec:retrieval} to save computing resources.

Due to the cohesive nature of knowledge, $\mathcal{U}_q^m$ can naturally be divided into $p$ connected components, denoted by $\mathcal{B}_i,1\leq i\leq p$, where nodes within each connected component $\mathcal{B}_i$ represent entities from the KG. Note that multiple edges may connect a pair of nodes due to redundant knowledge, which promotes us to generate the maximum spanning tree (MST) of each connected component for filtering. This can be formulated as
\begin{equation}
\label{eq:mst}
\mathcal{T}_i = \text{MST}(\mathcal{B}_i).
\end{equation}
Through such a filtering process, we retain only the most relevant linking information between entities and eliminate redundant edges, thereby enhancing the informativeness of the retrieved chunks.

\paragraph{Serving as an Arranger}
With the KG-based context organization module, we aim to integrate the retrieved chunks into intrinsically related and self-consistent paragraphs with the KG as the skeleton.

To achieve this, we provide two representations for each generated MST $\mathcal{T}_i$, including a text representation and a triplet representation. For the text representation, we pick the edge with the highest weight as the root, and concatenate all the chunks linked to the edges using a depth-first search (DFS) algorithm to form a coherent paragraph. For the triplet representation, we concatenate all the edges in the form of $<h,r,t>$ within the MST.

We calculate the relevance scores between MSTs and the user query based on their triplet representations using a cross-encoder reranking function~\cite{xiao23cpack}:
\begin{equation}
\label{eq:rerank}
R(q,\mathcal{T}_i)=C(q,\text{conc}(\mathcal{T}_i)),
\end{equation}
where $C(\cdot)$ is the cross-encoder reranking function and $\text{conc}(\cdot)$ is used to obtain the triplet representations. We use triplet representations instead of text representations because triplets provide a concise and structured refinement of the key information associated with the corresponding chunks, allowing relevance matching to focus on key information.

After computing the relevance scores, we sort the MSTs $\{\mathcal{T}_i\,|\,1\leq i\leq p\}$ according to their relevance $\{R(q,\mathcal{T}_i)\}$ to the user query $q$ in descending order. Then, we include their text representations in order until the top-$k$ constraint on the number of chunks has been reached. Finally, these selected chunks are fed into the LLMs along with the user query for response generation.

\begin{table*}[!ht]
\centering
\resizebox{\linewidth}{!}{
\begin{tabular}{lcccccccccccc}
\toprule
\multirowcell{2.4}{Methods} & \multicolumn{3}{c}{Hotpot-Dist} & \multicolumn{3}{c}{Hotpot-Full} & \multicolumn{3}{c}{Shuffle-Hotpot-Dist} & \multicolumn{3}{c}{Shuffle-Hotpot-Full} \\
\cmidrule(lr){2-4}\cmidrule(lr){5-7}\cmidrule(lr){8-10}\cmidrule(lr){11-13}
& F1 & Precision & Recall & F1 & Precision & Recall & F1 & Precision & Recall & F1 & Precision & Recall \\
\midrule
LLM-only & 0.237 & 0.259 & 0.234 & 0.237 & 0.259 & 0.234 & 0.158 & 0.175 & 0.158 & 0.158 & 0.175 & 0.158 \\
Semantic RAG & 0.617 & 0.646 & 0.643 & 0.528 & 0.558 & 0.535 & 0.508 & 0.533 & 0.524 & 0.422 & 0.449 & 0.433 \\
\quad +\,Rerank & 0.652 & 0.685 & 0.665 & 0.587 & 0.613 & 0.603 & 0.532 & 0.560 & 0.546 & 0.447 & 0.476 & 0.456 \\
Hybrid RAG & 0.653 & 0.676 & 0.655 & 0.551 & 0.582 & 0.558 & 0.520 & 0.548 & 0.534 & 0.443 & 0.473 & 0.446 \\
LightRAG & 0.293 & 0.288 & 0.480 & 0.261 & 0.259 & 0.364 & 0.285 & 0.284 & 0.404 & 0.202 & 0.199 & 0.293 \\
GraphRAG & 0.400 & 0.408 & 0.491 & 0.169 & 0.157 & 0.429 & 0.351 & 0.365 & 0.401 & 0.163 & 0.155 & 0.362 \\
\modelname & \textbf{0.663} & \textbf{0.690} & \textbf{0.683} & \textbf{0.631} & \textbf{0.665} & \textbf{0.643} & \textbf{0.545} & \textbf{0.572} & \textbf{0.566} & \textbf{0.507} & \textbf{0.539} & \textbf{0.512} \\
\bottomrule
\end{tabular}
}
\caption{Comparisons in terms of response quality between \modelname and baselines.}
\label{tab:results_answer}
\end{table*}

\begin{table*}[!ht]
\centering
\resizebox{\linewidth}{!}{
\begin{tabular}{lcccccccccccc}
\toprule
\multirowcell{2.4}{Methods} & \multicolumn{3}{c}{Hotpot-Dist} & \multicolumn{3}{c}{Hotpot-Full} & \multicolumn{3}{c}{Shuffle-Hotpot-Dist} & \multicolumn{3}{c}{Shuffle-Hotpot-Full} \\
\cmidrule(lr){2-4}\cmidrule(lr){5-7}\cmidrule(lr){8-10}\cmidrule(lr){11-13}
& F1 & Precision & Recall & F1 & Precision & Recall & F1 & Precision & Recall & F1 & Precision & Recall \\
\midrule
Semantic RAG & 0.343 & 0.206 & 0.894 & 0.300 & 0.178 & 0.790 & 0.321 & 0.201 & 0.837 & 0.268 & 0.167 & 0.708 \\
\quad +\,Rerank & 0.357 & 0.224 & \textbf{0.932} & 0.306 & 0.197 & 0.833 & 0339 & 0.213 & \textbf{0.886} & 0.286 & 0.179 & 0.754 \\
Hybrid RAG & 0.354 & 0.222 & 0.921 & 0.302 & 0.189 & 0.795 & 0.334 & 0.210 & 0.837 & 0.279 & 0.174 & 0.739 \\
LightRAG & 0.234 & 0.150 & 0.638 & 0.132 & 0.083 & 0.340 & 0.227 & 0.148 & 0.535 & 0.116 & 0.073 & 0.295 \\
GraphRAG & 0.255 & 0.167 & 0.594 & 0.180 & 0.113 & 0.470 & 0.210 & 0.138 & 0.482 & 0.199 & 0.126 & 0.510 \\
\modelname & \textbf{0.436} & \textbf{0.301} & 0.908 & \textbf{0.310} & \textbf{0.203} & \textbf{0.838} & \textbf{0.405} & \textbf{0.279} & 0.840 & \textbf{0.305} & \textbf{0.193} & \textbf{0.790} \\
\bottomrule
\end{tabular}
}
\caption{Comparisons in terms of retrieval quality between \modelname and baselines.}
\label{tab:results_sps}
\end{table*}

\section{Experiments}

\subsection{Experiment Setup}

\begin{figure}
\centering
\includegraphics[width=\linewidth]{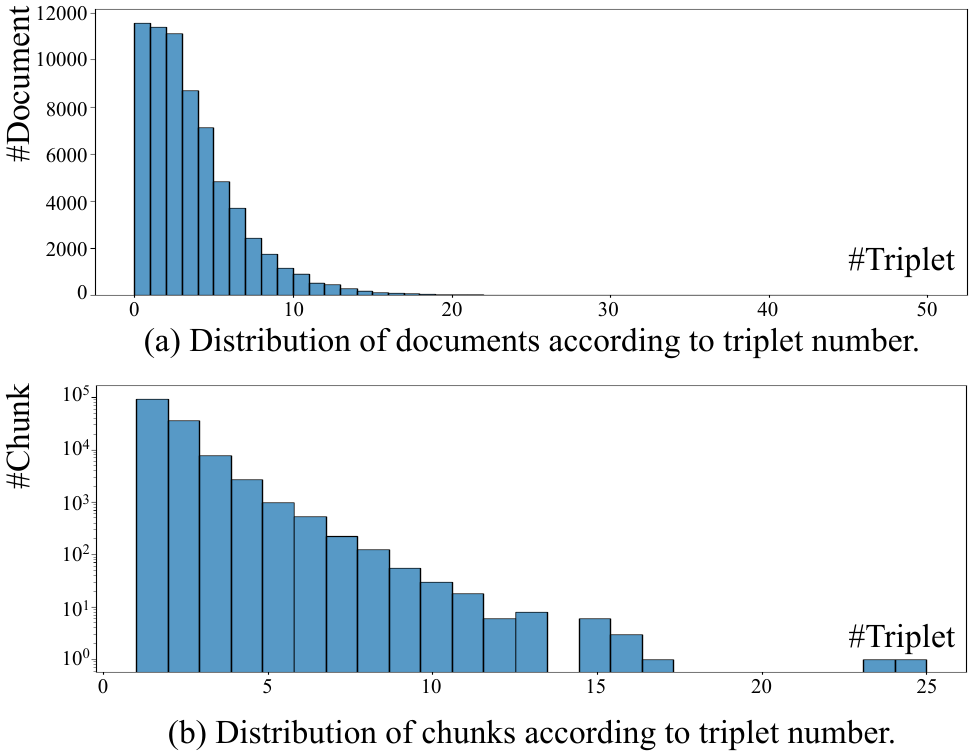}
\caption{Statistics of triplet extraction.}
\label{fig:triplet_distribution}
\end{figure}

\paragraph{Datasets}
We conduct experiments on the benchmark dataset HotpotQA~\cite{yang18hotpotqa}, where each query can be associated with several materials (e.g., relevant content in Wikipedia) to help in response generation. 
The HotpotQA dataset consists of two settings, named {\bf HotpotQA-Dist} and {\bf HotpotQA-Full}. In the distractor setting, a total of ten documents are provided as supporting materials, including all useful knowledge as well as some irrelevant content. In the fullwiki setting, it is required to identify useful knowledge from the entire 66,581 documents extracted from Wikipedia.

For the KG-chunk association, we provide a manual prompt to Llama-3~\cite{dubey24llama3} for extracting entities and relations from the 66,581 documents of HotpotQA, resulting in a total of 211,356 triplets consisting of 98,226 entities and 19,813 relations. Each triplet in the constructed KG is linked to its source chunk. 
We record the number of triplets extracted from each chunk and document, and plot the corresponding distributions of chunks and documents in Fig.~\ref{fig:triplet_distribution}, which shows a long-tail phenomenon.

Furthermore, to alleviate the dependence on prior knowledge during the generation process (i.e., the training corpus of LLMs may contain Wikipedia content) and to better demonstrate the effects of RAG, we construct variants of HotpotQA. Specifically, for each entity, we randomly replace it with another entity in the same category, and then update the queries, triplets, and documents accordingly. For example, the entity \textit{Family Guy} can be replaced with \textit{Rick and Morty}, and all instances of \textit{Family Guy} contained in queries, triplets, and documents would be updated to \textit{Rick and Morty}. Therefore, LLMs have to identify and extract relevant content from the documents rather than relying on prior knowledge about \textit{Family Guy} from training data to correctly answer the queries.
Note there might generate lots of new triplets such as (\textit{Rick and Morty}, \textit{language}, \textit{French}), as the original tail entity can be also transformed from \textit{English} to \textit{French}.
The produced variant datasets are denoted by {\bf Shuffle-HotpotQA-Dist} and {\bf Shuffle-HotpotQA-Full}, respectively.

\paragraph{Evaluation Metrics}
We compare \modelname with existing RAG-based methods in terms of response quality and retrieval quality, which can be influenced by both the retrieved chunks and context organization.
For retrieval quality, we use the evaluation script provided by HotpotQA to measure the F1 score, precision, and recall between the retrieved chunks and referenced facts.
For response quality, we adopt the F1 score, precision, and recall as metrics, comparing the generated responses against ground truth answers.

\paragraph{Baselines}
In the experiments, we compare \modelname with the following baseline methods:
\begin{itemize}
    \item \textit{LLM-only}, which directly instructs LLMs to generate responses to user queries without any additional retrieval mechanisms.
    \item \textit{Semantic RAG}~\cite{jiang23arag}, which employs a semantic-based approach to retrieve relevant chunks. These chunks are concatenated into the prompt and fed into the LLMs for response generation. For more details, please refer to Sec.~\ref{subsec:retrieval}.
    \item \textit{Hybrid RAG}~\cite{gao21complement}, which combines a semantic-based retrieval method with a keyword-based retrieval method (e.g., BM25~\cite{arian23bm25}) for chunk retrieval. The retrieved chunks are subsequently merged through a cross-encoder reranker.
    \item \textit{GraphRAG}~\cite{darren24graphrag}, which constructs a graph-based index with an LLM. GraphRAG derives a knowledge graph from the source documents and pre-generates community summaries for clustered entities. Given a query, it generates partial responses with each related community summary and aggregates them into the final answer.
    \item \textit{LightRAG}~\cite{guo2024lightrag}, which acts as a lightweight version of GraphRAG. LightRAG extracts entities and relations from the source documents and generates a short description of each entity for retrieval. The retrieved information is unified with the query and fed into the LLM for generation.
\end{itemize}

For \modelname and all baseline methods, we use LLaMA3-8B~\cite{dubey24llama3} as the LLM for KG construction and response generation, mxbai-embed-large~\cite{li24mxbai} as the embedding model, and bge-reranker-large~\cite{xiao23cpack} as the cross-encoder reranker for both Hybrid RAG and \modelname. The value of $k$ is set to 10 unless otherwise specified.

\begin{table*}[!ht]
\centering
{\small
\begin{tabular}{lccccccr}
\toprule
\multirowcell{2.4}{} & \multicolumn{3}{c}{Response Quality} & \multicolumn{4}{c}{Retrieval Quality} \\
\cmidrule(lr){2-4}\cmidrule(lr){5-8}
& F1 & Precision & Recall & F1 & Precision & Recall & $\#$Avg.\\
\midrule
\modelname & 0.663 & 0.690 & 0.683 & 0.436 & 0.301 & 0.908  & 8.11 \\
\quad w/o organization & 0.660 & 0.678 & 0.679 & 0.259 & 0.153 & 0.963 & 16.76 \\
\quad w/o expansion & 0.626 & 0.653 & 0.645 & 0.473 & 0.341 & 0.842 & 4.41 \\
\bottomrule
\end{tabular}
}
\caption{Experimental results of an ablation study conducted on HotpotQA in the distractor setting.}
\label{tab:raw_distractor_ablation}
\end{table*}

\begin{table*}[!ht]
\centering
{\small
\begin{tabular}{lccccccr}
\toprule
\multirowcell{2.4}{} & \multicolumn{3}{c}{Response Quality} & \multicolumn{4}{c}{Retrieval Quality} \\
\cmidrule(lr){2-4}\cmidrule(lr){5-8}
& F1 & Precision & Recall & F1 & Precision & Recall & $\#$Avg.\\
\midrule
\modelname & 0.545 & 0.572 & 0.566 & 0.405 & 0.279 & 0.840 & 8.09 \\
\quad w/o organization & 0.538 & 0.563 & 0.560 & 0.182 & 0.102 & 0.962 & 24.56 \\
\quad w/o expansion & 0.474 & 0.503 & 0.485 & 0.511 & 0.458 & 0.656 & 3.82 \\
\bottomrule
\end{tabular}
}
\caption{Experimental results of an ablation study conducted on Shuffle-HotpotQA in the distractor setting.}
\label{tab:pu_distractor_ablation}
\end{table*}

\begin{figure*}[t]
\centering
\includegraphics[width=0.85\linewidth]{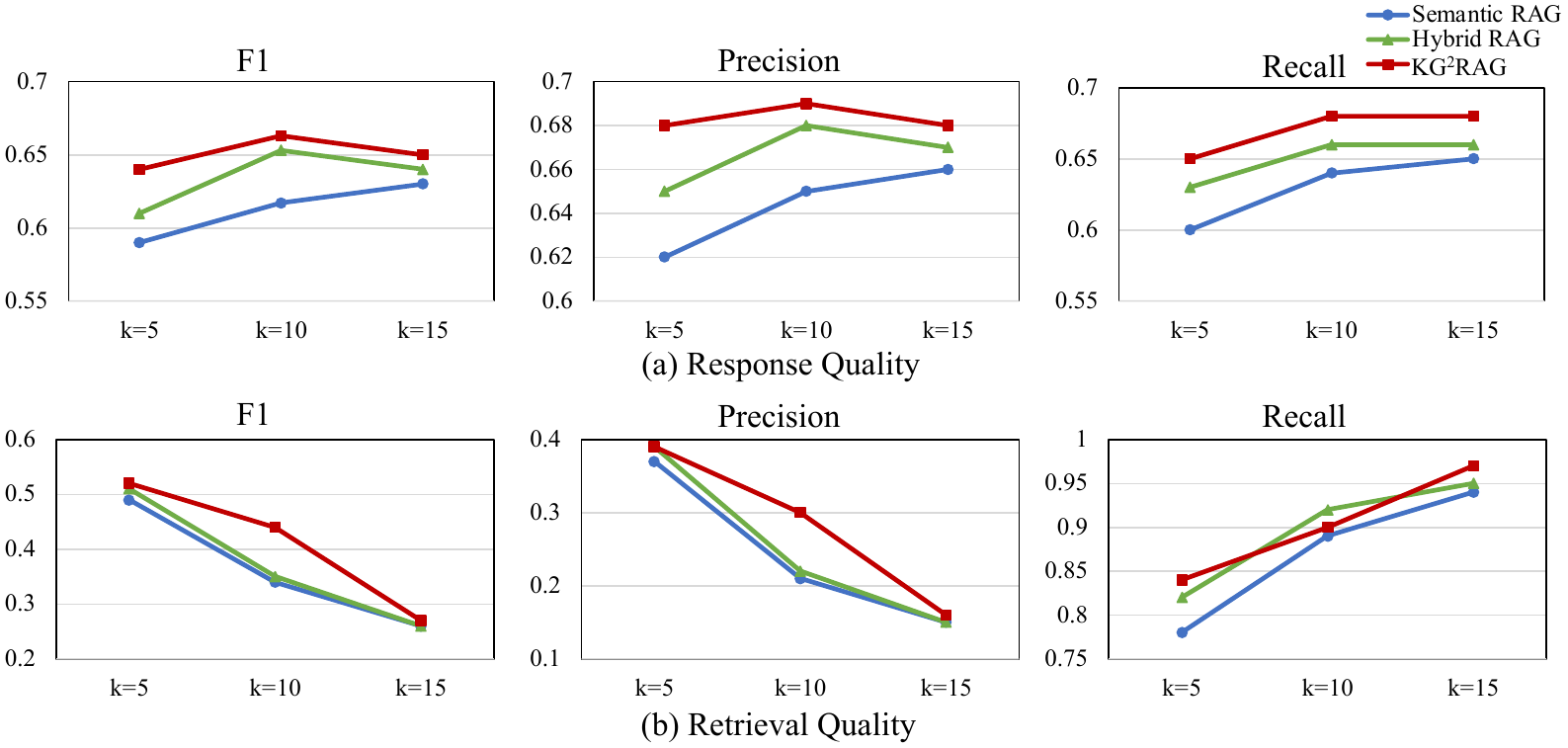}
\caption{Experimental results with varying top-$k$ on HotpotQA in distractor setting.}
\label{fig:vary_k}
\end{figure*}

\begin{table*}[t]
\centering
{\small
\begin{tabular}{lccccccr}
\toprule
\multirowcell{2.4}{} & \multicolumn{3}{c}{Response Quality} & \multicolumn{4}{c}{Retrieval Quality} \\
\cmidrule(lr){2-4}\cmidrule(lr){5-8}
& F1 & Precision & Recall & F1 & Precision & Recall & $\#$Avg.\\
\midrule
$m=1$ & 0.663 & 0.690 & 0.683 & 0.436 & 0.301 & 0.908 & 8.11 \\
$m=2$ & 0.656 & 0.681 & 0.674 & 0.420 & 0.291 & 0.917 & 8.53 \\
$m=3$ & 0.658 & 0.678 & 0.675 & 0.421 & 0.284 & 0.924 & 8.19 \\
\bottomrule
\end{tabular}
}
\caption{Experimental results on HotpotQA in the distractor setting with varying $m$.}
\label{tab:vary_m}
\end{table*}

\begin{table*}[t]
\centering
{\small
\begin{tabular}{lcccccc}
\toprule
\multirowcell{2.4}{} & \multicolumn{3}{c}{Response Quality} & \multicolumn{3}{c}{Retrieval Quality} \\
\cmidrule(lr){2-4}\cmidrule(lr){5-7}
& F1 & Precision & Recall & F1 & Precision & Recall \\
\midrule
Hybrid RAG & 0.653 & 0.676 & 0.655 & 0.354 & 0.222 & 0.921 \\
\modelname & 0.663 & 0.690 & 0.683 & 0.436 & 0.301 & 0.908 \\
\quad $-5\%$ & 0.662 & 0.681 & 0.676 & 0.434 & 0.306 & 0.898 \\
\quad $-10\%$ & 0.654 & 0.688 & 0.682 & 0.432 & 0.305 & 0.890 \\
\bottomrule
\end{tabular}
}
\caption{Experimental results on HotpotQA in the distractor setting with triplets dropped.}
\label{tab:drop_triplets}
\end{table*}

\subsection{Comparisons and Analyses}
\label{sec:comparison}

\paragraph{Response Quality}
The comparisons in terms of response quality between \modelname and the baselines are shown in Table~\ref{tab:results_answer}. From the table, we can observe that all methods utilizing RAG achieve significant improvements compared to the LLM-only approach, exceeding 29.1\% improvements in F1 scores on the original HotpotQA and 26.4\% improvements in F1 scores on Shuffle-HotpotQA. Among these RAG-based methods, \modelname achieves consistent outperformance, especially in the fullwiki setting and on the Shuffle-HotpotQA dataset.

In the fullwiki setting, a large pool of candidate documents (thousands of times more than in the distractor setting) is provided to LLMs, necessitating high-quality retrieval results and effective context organization. In such a challenging setup, our proposed method \modelname achieves at least 8\% improvements compared to baselines, demonstrating that \modelname enhances chunk retrieval through KG-guided approaches that surpass semantic-based and keyword-based methods. Besides, on the Shuffle-HotpotQA dataset, where LLMs should rely more on RAG rather than prior knowledge, our proposed method achieves at least 2.5\% and 6.4\% improvements in the distractor and fullwiki settings, respectively.

\paragraph{Retrieval Quality}
The experimental results are shown in Table~\ref{tab:results_sps}, which demonstrate that \modelname strikes a favorable balance between retrieval precision and recall, highlighting the effectiveness of KG-guided expansion and context organization. In the distractor setting, where irrelevant chunks are limited, our proposed method achieves similar performance in recall but significantly better performance in precision (more than 7.9\% and 6.9\% on HotpotQA and Shuffle-HotpotQA, respectively). 
In the fullwiki setting where identifying relevant chunks is more challenging, our proposed method achieves consistent improvements in both precision and recall compared to other RAG-based methods. These results further confirm the effectiveness of \modelname in providing high-quality retrieval results with the help of KG.

\subsection{Further Discussions}
\label{sec:discussions}

\paragraph{Ablation Study}
We conduct an ablation study to demonstrate the contributions of different modules in \modelname, including KG-guided expansion and KG-based context organization. The experimental results on the HotpotQA and Shuffle-HotpotQA datasets in the distractor setting are shown in Tables~\ref{tab:raw_distractor_ablation} and \ref{tab:pu_distractor_ablation}, where we also report the average number of retrieved chunks.

From these results, we can observe that using only KG-guided expansion without KG-based context organization (denoted by ``w/o organization'' in the table), \modelname achieves similar performance in terms of answer quality but significantly worse retrieval quality. The reason is that, without the KG-based context organization module, the number of retrieved chunks can be noticeably larger, potentially containing irrelevant chunks that do not contribute positively to performance but consume additional tokens. These findings confirm the contribution of the KG-based context organization module in effectively selecting and organizing retrieved chunks to preserve relevant information. 

With only the KG-based context organization module (denoted by ``w/o expansion'' in the table), \modelname achieves high retrieval precision and F1 score with a significantly smaller number of chunks, but fails to provide better responses, as some necessary chunks may not be retrieved using only semantic-based approaches.
These results confirm the importance of the KG-guided expansion module in successfully leveraging KG to capture fact-level relationships between chunks and retrieve key information that might be missed by semantic-based approaches.

\paragraph{Performance w.r.t. Varying $k$}
We conduct experiments with varying top-$k$ values on HotpotQA in the distractor setting. The experimental results are shown in Fig.~\ref{fig:vary_k}. 
From these figures, we can observe that \modelname maintains superior performance compared to baselines with different $k$. When $k$ is set to a suitable value (e.g., 5 or 10), \modelname ensures the efficient retrieval of high-quality chunks, thereby providing coherent and contextually consistent contexts for generating high-quality responses.

However, when $k$ is set to a too large value (e.g., 15), although the retrieval recall significantly improves, the quality of the generated responses does not increase proportionally, which indicates simply increasing the number of chunks cannot always result in a better retrieval recall ratio and response quality. 
\modelname exhibits the least sensitivity to the hyperparameter $k$ compared to baselines, which makes the RAG process robust.

\paragraph{Performance w.r.t. Varying $m$}
In \modelname, $m$ serves as the hyperparameter for graph expansion, balancing the trade-off between retrieval precision and recall. We set the $m$-hop value to 1 in the previous experiments. 
To further explore the effects of $m$, we conduct experiments with varying $m$ on HotpotQA dataset. The results are shown in Table~\ref{tab:vary_m}. These results indicate that setting $m=1$ is appropriate for the experiments, and KG$^2$RAG shows low sensitivity to the hyperparameter $m$.

\paragraph{Robustness Analysis}
To further confirm the robustness of \modelname with quality-limited KGs, we randomly drop 5\% or 10\% of the triplets from the constructed KG, and show the experimental results in Table~\ref{tab:drop_triplets}. The results demonstrate that \modelname maintains robust performance even with quality limitations and outperforms the baselines.
\section{Related Work}

\paragraph{Retrieval-augmented Generation}
To address the issues of hallucinations~\cite{xu24hallucinationsurvey,liu24hallucinationsurvey} due to a lack of corresponding knowledge or containing outdated knowledge, retrieval-augmented generation (RAG)~\cite{gao23ragsurvey,fan24ragsurvey} has been proposed for retrieving relevant chunks from a pool of candidate documents to assist LLM generation.

In a typical RAG system~\cite{patrick20rag}, the documents are first segmented into chunks based on lengths and structures, and then encoded with an embedding model~\cite{zach24nomic,li24mxbai} and indexed for efficient retrieval.
Inspired by the idea of sliding windows~\cite{jiao06slidingwindows}, sentence window retrieval~\cite{jiang23arag,matous24aragog} fetches the neighboring chunks around the retrieved chunks and concatenates them into a single larger chunk for context enrichment.
However, sentence window retrieval only considers the physical proximity of text chunks within the same document.
Different from existing studies, \modelname performs retrieval expansion based on factual associations among chunks that may be across multiple documents.

Reranking~\cite{nicholas24improving,michael22re2g} is a critical technique in information retrieval~\cite{mandalay62irsurvey,kuo24irsurvey}. In RAG systems, feeding the retrieved chunks along with the queries into a deep learning-based cross-encoder~\cite{xiao23cpack} can measure the semantic relevance more precisely, thereby enhancing both the retrieval and generation quality.
\modelname organizes the retrieved chunks into paragraphs with KGs as the skeleton, allowing a fine-grained measurement of paragraph-level relevance to queries.

\paragraph{LLMs with Knowledge Graph}
LLM~\cite{li24llmsurvey,ren24llmsurvey} is one of the most representative achievements of contemporary artificial intelligence (AI).
KGs~\cite{ji22kgsurvey}, as graph-structured relational databases, serve as a crucial data infrastructure for AI applications.
Research indicates that LLMs have the potential to address tasks related to KGs, such as knowledge graph completion~\cite{liu24finetuning} and knowledge graph question answering~\cite{sen23knowledge}.

Recently, the research community begins to explore how KGs can be used to enhance the generation capability of LLMs~\cite{wang24kgp,darren24graphrag,xu24retrieval}.
For example,
KGP~\cite{wang24kgp} constructs a document KG consisting of page and passage nodes, and links passage nodes with TF-IDF. The document KG is employed for retrieval expansion. The document KG constructed by KGP is based on sentence-level text similarity, which essentially functions similarly to simply expanding the context window.
GraphRAG~\cite{darren24graphrag} targets at query-focused summarization tasks. GraphRAG extracts KGs automatically from the document base with an LLM and analyzes the semantic structure of the dataset before querying, by splitting the KG from different level and detecting linked nodes hierarchically.
Different from previous studies, \modelname aims to enhance RAG with the fact-level structure and factual knowledge of KGs.

\section{Conclusion}
In this paper, we propose \modelname, a novel framework designed to enhance the performance of RAG through the integration of KGs. We introduce linkages between chunks and a specific KG, which help in providing fact-level relationships among these chunks. Consequently, \modelname suggests performing the KG-guided chunk expansion and the KG-based context organization based on seed chunks retrieved by semantic-based retrieval approaches.
Through these processes, the retrieved chunks become diverse, intrinsically related, and self-consistent, forming well-organized paragraphs that can be fed into LLMs for high-quality response generation. We compare \modelname with existing RAG-based approaches, demonstrating its superior performance in both response quality and retrieval quality. An ablation study is also conducted to further confirm the contributions of KG-guided chunk expansion and KG-based context organization, indicating that these two modules collaboratively enhance the effectiveness of \modelname.

\section*{Acknowledgments}
This work is supported by the National Natural Science Foundation of China (No. 62272219).

\section*{Limitations}

Retrieval-augmented generation (RAG) is a systematic engineering framework that can be refined from multiple perspectives, including query rewriting ~\cite{xiao23cpack}, retrieval optimization ~\cite{matous24aragog}, multi-turn dialogue ~\cite{Yao23iclr} and so on \cite{gao23ragsurvey}. 
\modelname only focuses on the part of retrieval optimization and aims to perform KG-guided retrieval expansion and KG-based context organization to enhance RAG with the structured factual knowledge from KGs, without optimizing other modules.
However, the proposed \modelname is orthogonal and compatible with the aforementioned modules. 
In the future, we will develop \modelname into a plug-and-play tool that can be easily integrated with other approaches, thereby better facilitating the research community.

\bibliography{custom}

\clearpage
\appendix

\section{Additional Experimental Results}

\subsection{Results on More Datasets}
\label{app:more_datasets}
We conduct additional experiments on two different datasets to confirm the effectiveness and generality of \modelname in various scenarios.

As shown in Table~\ref{tab:results_musique}, \modelname maintains superiority on the widely-used MuSiQue dataset~\cite{harsh22musique} in response F1 score, response exact match (EM) rate, and retrieval F1 score.

\begin{table}[!ht]
\centering
\resizebox{\columnwidth}{!}{
\begin{tabular}{lccc}
\toprule
Methods & Response F1 & Response EM & Retrieval F1 \\
\midrule
LLM-only & 0.075 & 0.025 & - \\
Semantic RAG & 0.367 & 0.248 & 0.365 \\
\quad +\,Rerank & 0.380 & 0.249 & 0.372 \\
Hybrid RAG & 0.380 & 0.250 & 0.364 \\
LightRAG & 0.248 & 0.170 & 0.289 \\
GraphRAG & 0.231 & 0.156 & 0.273 \\
\modelname & \textbf{0.419} & \textbf{0.303} & \textbf{0.451} \\
\bottomrule
\end{tabular}
}
\caption{Comparison results on MuSiQue.}
\label{tab:results_musique}
\end{table}

Also, we conduct experiments on the typical long-context dataset TriviaQA~\cite{joshi17triviaqa}. On average, each document in this dataset contains 2,895 words. For comparison, documents in HotpotQA have an average of 917 words. The experimental results are shown in Table~\ref{tab:results_trivia}, which confirms the effectiveness of \modelname in a typical long-context setting.

\begin{table}[!ht]
\centering
\resizebox{\columnwidth}{!}{
\begin{tabular}{lccc}
\toprule
Methods & Response F1 & Response Prec. & Response Recall \\
\midrule
LLM-only & 0.182 & 0.303 & 0.144 \\
Semantic RAG & 0.259 & 0.413 & 0.211 \\
\quad +\,Rerank & 0.265 & 0.409 & 0.235 \\
Hybrid RAG & 0.262 & 0.415 & 0.229 \\
LightRAG & 0.118 & 0.157 & 0.237 \\
GraphRAG & 0.127 & 0.193 & 0.225 \\
\modelname & \textbf{0.273} & \textbf{0.416} & \textbf{0.240} \\
\bottomrule
\end{tabular}
}
\caption{Comparison results on TriviaQA.}
\label{tab:results_trivia}
\end{table}

\subsection{Efficiency Analysis}
\label{app:efficiency_analysis}
We compare the KG construction cost of \modelname with two other KG-enhanced RAG approaches: LightRAG~\cite{guo2024lightrag} and GraphRAG~\cite{darren24graphrag}. The results, as summarized in Table~\ref{tab:kg_efficiency}, demonstrate that \modelname is more efficient in terms of token cost, the number of LLM calls, and time cost.

\begin{table}[!ht]
\centering
\resizebox{\columnwidth}{!}{
\begin{tabular}{l|cccc}
\toprule
& $\#$Input tokens & $\#$Output tokens & $\#$LLM calls & Extraction time \\
\midrule
LightRAG & 1,269 & 381 & 1 & 3s \\
GraphRAG & 2,791 & 629 & 5 & 6s \\
\modelname & \ \ \ 561 & \ \ 22 & 1 & 1s \\
\bottomrule
\end{tabular}
}
\caption{Comparison of average LLM and time cost per chunk during KG construction.}
\label{tab:kg_efficiency}
\end{table}

We calculate the average retrieval time and generation time of \modelname compared to LightRAG and GraphRAG. The results in Table~\ref{tab:rag_efficiency} indicate that \modelname requires less time for both retrieval and response generation than LightRAG and GraphRAG, and is very close to Semantic RAG. Note that \modelname might need a lower time for response generation using a condensed and informative context as input.

\begin{table}[!ht]
\centering
\resizebox{\columnwidth}{!}{
\begin{tabular}{l|cc}
\toprule
Method & Avg. retrieval time & Avg. generation time \\
\midrule
Semantic RAG & 21ms & 2,500ms \\
LightRAG & 40ms & 5,600ms \\
GraphRAG & 42ms & 5,500ms \\
\modelname & 25ms & 2,300ms \\
\bottomrule
\end{tabular}
}
\caption{Comparison of average retrieval and generation time per query.}
\label{tab:rag_efficiency}
\end{table}

\end{document}